\documentclass[format=sigconf, review=false, screen=true]{acmart}

\usepackage{booktabs} 
\usepackage{subfigure}

\usepackage[ruled]{algorithm2e} 

\SetAlFnt{\small}
\SetAlCapFnt{\small}
\SetAlCapNameFnt{\small}
\SetAlCapHSkip{0pt}
\IncMargin{-\parindent}

\acmJournal{TWEB}
\acmVolume{9}
\acmNumber{4}
\acmArticle{39}
\acmYear{2010}
\acmMonth{3}
\copyrightyear{2009}

\setcopyright{rightsretained}

\acmDOI{0000001.0000001}

\begin{document}
\title[Bike Flow Prediction with Multi-Graph Convolutional Networks]{Bike Flow Prediction with Multi-Graph Convolutional Networks}
\author{Di Chai$^*$, Leye Wang$^*$, Qiang Yang \\
Hong Kong University of Science and Technology, Hong Kong SAR, China \\
\large
dchai@connect.ust.hk, wangleye@gmail.com, qyang@cse.ust.hk\\
\large ($^*$Equal contribution, ordered alphabetically)}
\begin{abstract} One fundamental issue in managing bike sharing systems is the bike flow prediction. Due to the hardness of predicting the flow for a single station, recent research works often predict the bike flow at cluster-level. While such studies gain satisfactory prediction accuracy, they cannot directly guide some fine-grained bike sharing system management issues at station-level. In this paper, we revisit the problem of the station-level bike flow prediction, aiming to boost the prediction accuracy leveraging the breakthroughs of deep learning techniques. We propose a new multi-graph convolutional neural network model to predict the bike flow at station-level, where the key novelty is viewing the bike sharing system from the graph perspective. More specifically, we construct multiple inter-station graphs for a bike sharing system. In each graph,  nodes are stations, and edges are a certain type of relations between stations. Then, multiple graphs are constructed to reflect heterogeneous relationships (e.g., distance, ride record correlation).  Afterward, we fuse the multiple graphs and then apply the convolutional layers on the fused graph to predict station-level future bike flow. In addition to the estimated bike flow value, our model also gives the prediction confidence interval so as to help the bike sharing system managers make decisions. Using New York City and Chicago bike sharing data for experiments, our model can outperform state-of-the-art station-level prediction models by reducing 25.1\% and 17.0\% of prediction error in New York City and Chicago, respectively.
\end{abstract}

%
%


%
%


\maketitle

\renewcommand{\shortauthors}{}

\section{Introduction}

Bike sharing systems are gaining increasing popularity in city transportation as a way to provide flexible transport mode and reduce the production of greenhouse gas. In a bike sharing system, users can check out at nearby stations and return the bike to the stations near the destination. Bike flow prediction is one of the key research and practical issues in bike sharing systems, which plays an important role in various tasks such as bike rebalancing~\cite{li2015traffic,chen2016dynamic}. 

In reality, the bike flow of a single station in a city usually has a very complicated dynamic pattern, which makes it hard to predict with traditional statistical learning or machine learning methods~\cite{chen2016dynamic}. As a result, most recent researchers try to address the bike flow prediction in a cluster-level. That is, they first group up the stations, and then predict the bike flow for each cluster~\cite{li2015traffic,chen2016dynamic}. Although the cluster-level prediction accuracy is more satisfied, there are two limitations: (i) whether the output clusters are appropriate or not is hard to evaluate as there is no ground truth, and (ii) the prediction result at cluster-level still cannot directly support the precise management on single stations. 

In this paper, we revisit the single station-level bike flow prediction problem in bike sharing systems, which can provide fine-grained information for the system administrators' decision making process and avoid the hard-to-evaluate clustering problem. To achieve this goal, we make effort from two aspects:

(i) \textbf{Propose a novel multi-graph convolutional neural network model to catch heterogeneous inter-station spatial correlations}: Traditional single station-level prediction usually pays more focus on the station's historical data, such as ARIMA \cite{williams2003modeling}. However, in addition to this temporal correlations, the inter-station spatial correlations may also play an important role in bike flow prediction. In this work, we propose a new multi-graph convolutional neural network to capture heterogeneous spatial relationships between stations, such as distance and historical usage correlations. After the multi-graph convolutional layers, an encoder-decoder structure including LSTM (Long-Short Term Memory) units~\cite{hochreiter1997long} is built to catch the temporal patterns. Hence, both spatial and temporal patterns are effectively captured for station-level bike flow prediction. 

(ii) \textbf{Compute confidence interval for the prediction}: The demand of single stations sometimes fluctuates a lot in reality. At that time, only providing the estimation value to the bike sharing system managers may not be enough. To this end, our model is designed to further compute the confidence of prediction to help managers make better decisions. More specifically, by leveraging the dropout techniques in neural networks, we simulate various realistic factors affecting the uncertainty of prediction, such as \textit{model uncertainty}, \textit{model misspecification}, and \textit{inherent noise}~\cite{zhu2017deep}. Based on these simulations, we can infer the confidence interval for our prediction accurately.

Briefly, this paper has the following contributions:

(i) To the best of our knowledge, this is the first work of leveraging graph convolutional neural networks in to predict station-level bike flow in a bike sharing system, as well as providing prediction confidence estimation.

(ii) We propose a novel \textit{multi-graph convolutional neural network} model to utilize heterogeneous spatial correlations between stations for station-level bike flow prediction. More specifically, three different graphs are constructed for a bike sharing system, i.e., \textit{distance}, \textit{interaction}, and \textit{correlation} graphs. A method to fuse multiple graphs is designed so that graph convolution can be applied on heterogeneous inter-station spatial correlations simultaneously. Then, an encoder-decoder structure with LSTM units is built to capture temporal patterns in historical bike flow records. By properly leveraging dropout techniques, our proposed model can not only calculate the bike flow prediction results, but also the confidence interval at the same time.

(iii) Evaluations on real bike flow dataset in New York City and Chicago shows the effectiveness of our mothod. Compared with the state-of-the-art station-level bike flow prediction models such as LSTM and ARIMA, our multi-graph convolutional neural network model can reduce up to 25.1\% prediction error.

\section{Related Work}

We describe the related work from two perspectives, \textit{bike flow prediction} and \textit{graph convolutional neural networks}.

\subsection{Bike Flow Prediction}

Flow prediction is a very important topic in bike sharing system. Current studies on bike flow prediction most fall into three categories which are \textit{cluster-based}, \textit{area-based}, and \textit{station-based} flow prediction. 

\textbf{Cluster-based flow prediction}: The demand of bike-sharing system is influenced by many factors such as weather, holiday, special events and the influence between stations. To make the prediciton result more accurate, Li et al. group the stations into several clusters using distance and bike usage information. Then they predicted the aggregate demand over all the stations and the proportion for each cluster \cite{li2015traffic}. Chen et al. used a graph based clustering method to get high internal connectivity, and then predict the over-demand probability of each cluster \cite{chen2016dynamic}. Cluster based flow prediction is also used in \cite{xu2013public}. Topics about building clusters in bike sharing system are also studied in \cite{vogel2011understanding, schuijbroek2017inventory, austwick2013structure, etienne2014model, zhou2015understanding}. While this steam of studies is very popular, the intrinsic difficulty for applying such techniques in real life is the cluster result may not be desired for bike sharing system administration. In most cases, providing station-level prediction is still more practical.

\textbf{Area-based flow prediction}: Unlike the cluster-based flow prediction, area-based methodology focuses on the bike flow of a specific area. One recent way is to conduct grid based map segmentation over the city, and then applies state-of-the-art deep models, such as CNN, ResNet, or ConvLSTM, to predict the flow of each area \cite{zhang2017deep, zhang2016dnn}. But this methodology does not work in single-station prediction because it is hard to decide the size of the area. More than one station will appear in one area if the grid size is large or many grids will contain no station if it is small. 

\textbf{Station-based flow prediction}: Compared with the first two types of flow prediction, station-based flow prediction is harder but can provide more fine-grained information in the system operation process. Jon Froehlich et al. compared four simple models in predicting available bikes' number which are \textit{last value}, \textit{historical mean}, \textit{historical trend} and \textit{Bayesian network} \cite{froehlich2009sensing}. Some researchers adopted time series analysis to predict the future bike demand \cite{vogel2011strategic, yoon2012cityride}. Kaltenbrunner et al. used a statistical model to predict the availability of bikes in the future \cite{kaltenbrunner2010urban}. Compared to these works, we are the first to apply deep learning methods for station-based bike flow prediction, and our experiments show that the performance improvement is significant.

\subsection{Graph Convolutional Neural Networks}

The graph convolutional neural network was first introduced by Bruna et al.~\cite{bruna2013spectral}, which applies the convolutional layers on the graph data. It is later extended by Defferrard et al.~\cite{NIPS2016_6081} with fast localized convolutions to accelerate the computation process. Kipf et al. proposed an efficient variant of convolutional neural network which can be used directly on graphs, and the network achieved good performance on graph node classification task \cite{kipf2016semi}. Seo et al. proposed a graph convolutional recurrent network which can deal with structured sequence data \cite{seo2016structured}. The implementation of graph convolutional network is also studied in image classification~\cite{yi2017syncspeccnn} and the  segmentation of point cloud~\cite{wang2018dynamic, zhou2016fine}. Two most relevant papers to our work are~\cite{li2018diffusion,yu2018spatio}, both applying graph convolutional neural networks to predict traffic speed in road segments. Our work is distinct from them in two aspects. First, \cite{li2018diffusion,yu2018spatio} only use distance to create a graph for road segments; however, as one graph may not be able to describe inter-station relationships comprehensively, we propose new ways (in addition to distance) to construct inter-station graphs and further design a multi-graph convolutional network structure. Second, our model can output prediction confidence interval, which can thus provide more information for the decision making process of bike sharing system organizers.

\section{Definitions and Problem}

In this section, we first define several key concepts, and then formulate the problem. 

\textbf{Definition 1}. \textit{Bike-Sharing System Graph}: The bike-sharing system is represented as a weighted graph, whose nodes are stations and edges are inter-station relationships. The weights  of edges represent the relation strength between stations. Usually, the larger weights mean that the two stations have higher correlations (e.g., the edge's weight can be the reciprocal of distance between two stations). How to construct the graph is one part of our method and will be elaborated in the next section.

\textbf{Definition 2}. \textit{Bike Flow}: There are two types of bike flow: inflow and outflow. Suppose we have $N$ bike stations, inflow at the time interval $t$ (e.g., one-hour) can be denoted as $I^t=[ci_1^t,ci_2^t,...,ci_N^t]$, outflow at the time interval $t$ can be denoted as $O^t=[co_1^t,co_2^t,...,co_N^t]$. 

\textbf{Problem}: Suppose the current time is $t-1$, and we have the history data $[(I^0,O^0),(I^1,O^1),...,(I^{t-1},O^{t-1})]$. The problem is to predict the bike flow at the next time $(\hat I^{t}, \hat O^{t})$, aiming to:
$$
\min ||\hat I^{t} - I^{t}||^2_2, \quad \min ||\hat O^{t} - O^{t}||^2_2
$$
where $(I^{t}, O^{t})$ is the ground truth bike flow of the next time $t$.

\section{Multi-Graph Convolutional Neural Network Model}

To solve the above problem, we propose a novel multi-graph convolutional neural network model, which will be elaborated next.

\subsection{Framework Overview}

\begin{Huge}
\begin{figure*}[h]
\centering
\includegraphics[width=.8\linewidth]{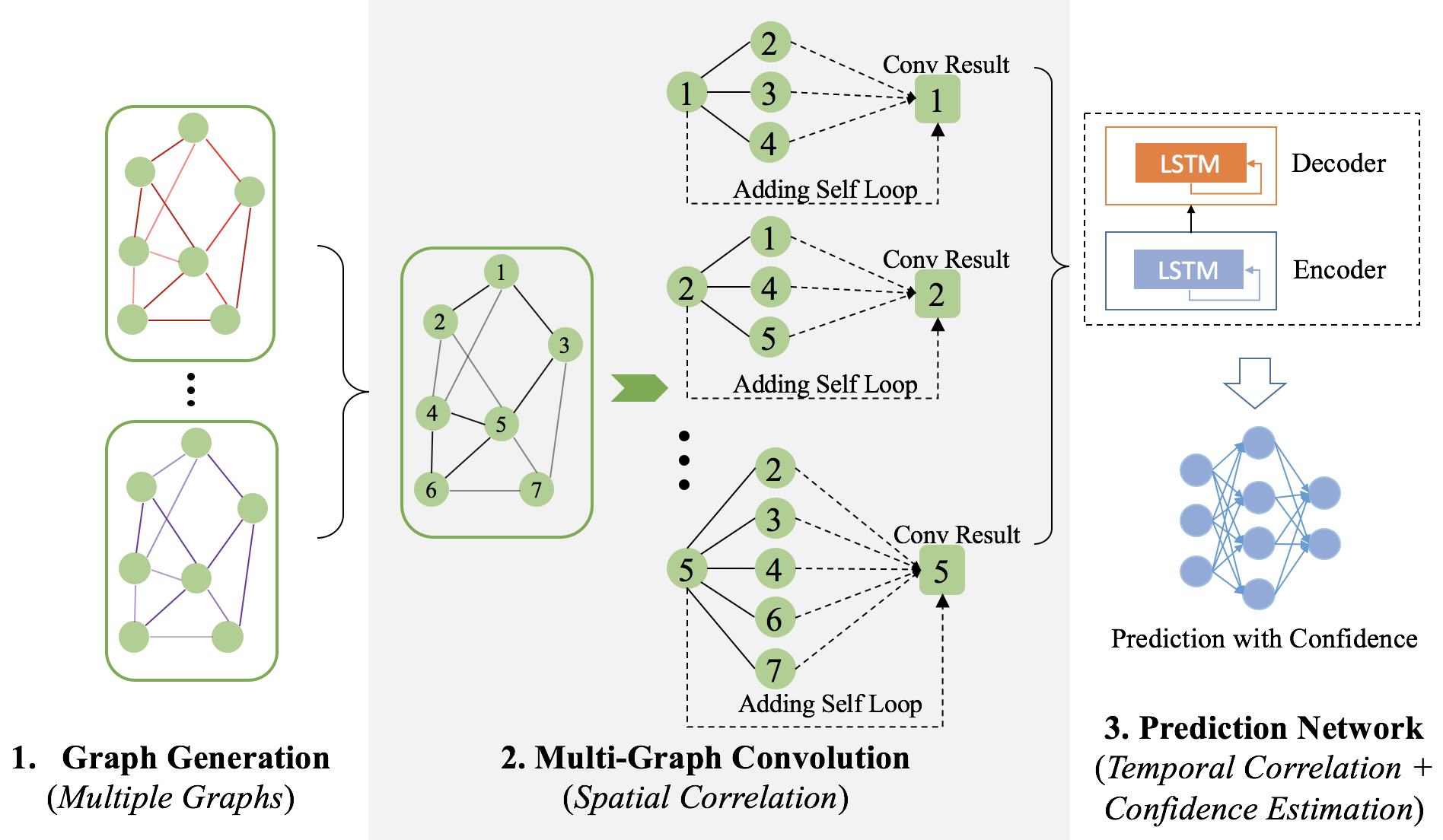}
\caption{Overview of the Multi-Graph Convolutional Neural Network Model for Bike Flow Prediction}
\label{fig:one}
\end{figure*}
\end{Huge}

Figure ~\ref{fig:one} gives an overview of our model containing three parts:

\textbf{(i) Graph Generation}: As the grid-based map segmentation~\cite{zhang2016dnn} does not work when we want to predict the single station's demand, we propose to build inter-station graphs to represent the bike-sharing system, where the links between stations reflect the spatial relationships. More specifically, the nodes in the graph are the bike stations, and the edges represent relationships between stations. We also encode weights on the edges as the relationship strength between stations can be different (e.g., smaller distance between stations may refer to a closer relationship). Moreover, since there may be various relationships between stations that can help our prediction, we construct multiple graphs, such as \textit{distance}, \textit{interaction}, and \textit{correlation}, which will be elaborated later.

\textbf{(ii) Multi-Graph Convolution}: As we construct multiple inter-station graphs to represent one bike sharing system, we introduce a multi-graph convolution part to perform the convolution operation considering all these graphs. More specifically, we first develop a fusion way to incorporate multiple graph information into one fused inter-station graph. Then, we use the graph convolutional layers on the fused graph to encode the graph structure and features of nodes \cite{kipf2016semi}.  In our fused bike-sharing graph, graph convolution can extract features of various \textit{spatial relationships} between stations.

\textbf{(iii) Prediction Network}: Based on the graph convolution result, the third step designs a  network structure to \textit{predict the bike flow and compute the confidence simultaneously}. More specifically, the first component of the prediction network is an encoder-decoder structure with LSTM (Long-Short Term Memory) units~\cite{hochreiter1997long}, which can learn the hidden representation for each station to catch the \textit{temporal correlations} in the bike flow history. Then, by decomposing the prediction uncertainty into three parts: \textit{model uncertainty}, \textit{model misspecification}, and \textit{inherent noise} \cite{zhu2017deep} (details will be elaborated later), we further estimate the confidence interval of our station-level bike flow prediction, which can provide more information to the managers of bike sharing systems for decision making. 

In next a few sections, we will elaborate each part of our multi-graph neural network model in more details.

\subsection{Detailed Solution}

\textbf{(i) Graph Generation}

Graph generation is the key to the success of graph convolutional model. If the constructed graph cannot encode the effective relationships between stations, it will not help the network parameter learning while even degrading the prediction performance. In general, we want to assign large weights to the edges between stations with similar dynamic flow patterns. Based on this idea, we propose three alternative ways for building inter-station graphs: \textit{distance graph}, \textit{interaction graph} and \textit{correlation graph}. 

\textbf{Distance Graph}: Tobler's first law of geography has pointed out that `\textit{everything is related to everything else, but near things are more related than distant things}'.\footnote{\url{https://en.wikipedia.org/wiki/Tobler\%27s_first_law_of_geography}} In bike sharing systems, for two stations near each other (e.g., around a metro station), they may share similar usage patterns. Following this idea, we use the distance to construct the inter-station graphs. More specifically, we use the reciprocal of the distance to mark the weight between two stations so that closer stations will be linked with higher weights.
$$G_d(V, E) \qquad weight = distance^{-1}$$
$$
A =
\left( \begin{array}{ccccc}
0&\frac{1}{dist_{0,1}}&\frac{1}{dist_{0,2}}&\cdots&\frac{1}{dist_{0,N-1}}\\
\frac{1}{dist_{1,0}}&0&\cdots&\cdots&\frac{1}{dist_{1,N-1}}\\
\frac{1}{dist_{2,0}}&\frac{1}{dist_{2,1}}&0&\cdots&\frac{1}{dist_{2,N-1}}\\
\vdots&\vdots&\vdots&\ddots&\vdots\\
\frac{1}{dist_{N-1,0}}&\cdots&\cdots&\cdots&0\\
\end{array} \right)
$$

\textbf{Interaction Graph}: The historical ride records can also provide plenty of information to construct the inter-station graphs. For example, if there exist many ride records between station $i$ and station $j$. Then the two stations $i$ and $j$ tend to affect each other regarding the dynamic bike flow patterns. With this idea in mind, we construct an \textit{interaction graph} to indicate whether two stations are interacted with each other frequently according to the historical ride records.  Denote $d_{i,j}$ as the number of ride records between $i$ and $j$, we build the interaction graph as:
$$G_i(V, E) \qquad weight=\# Riding Record Number$$
$$
A =
\left( \begin{array}{ccccc}
d_{0,0}&d_{0,1}&d_{0,2}&\cdots&d_{0,N-1}\\
d_{1,0}&d_{1,1}&d_{1,2}&\cdots&d_{1,N-1}\\
d_{2,0}&d_{2,1}&d_{2,2}&\cdots&d_{2,N-1}\\
\vdots&\vdots&\vdots&\ddots&\vdots\\
d_{N-1,0}&\cdots&\cdots&\cdots&d_{N-1,N-1}\\
\end{array} \right)
$$

\textbf{Correlation Graph}: With ride records, we also try another way to build the inter-station graph with the correlation of stations' historical usages. That is, we calculate the historical usages (inflow or outflow) of each station in each time slot (e.g., one hour), and then compute the correlations between every two stations as the inter-station link weights in the graph. In this work, we use the popular \textit{Pearson coefficient} to calculate the correlation. Denote $r_{i,j}$ as the Pearson correlation between station $i$ and station $j$, we can represent the correlation graph as follows:
$$G_c(V, E) \qquad weight=Correlation$$
$$r=\frac{\sum_{i=1}^n(X_i-\overline{X})(Y_i-\overline{Y})}{\sqrt{\sum_{i=1}^n(X_i-\overline{X})^2}\sqrt{\sum_{i=1}^n(Y_i-\overline{Y})^2}}$$
$$
A =
\left( \begin{array}{ccccc}
0&r_{0,1}&r_{0,2}&\cdots&r_{0,N-1}\\
r_{1,0}&0&r_{1,2}&\cdots&r_{1,N-1}\\
r_{2,0}&r_{2,1}&0&\cdots&r_{2,N-1}\\
\vdots&\vdots&\vdots&\ddots&\vdots\\
r_{N-1,0}&\cdots&\cdots&\cdots&0\\
\end{array} \right)
$$

\begin{figure}
	\centering
	\includegraphics[width=1\linewidth]{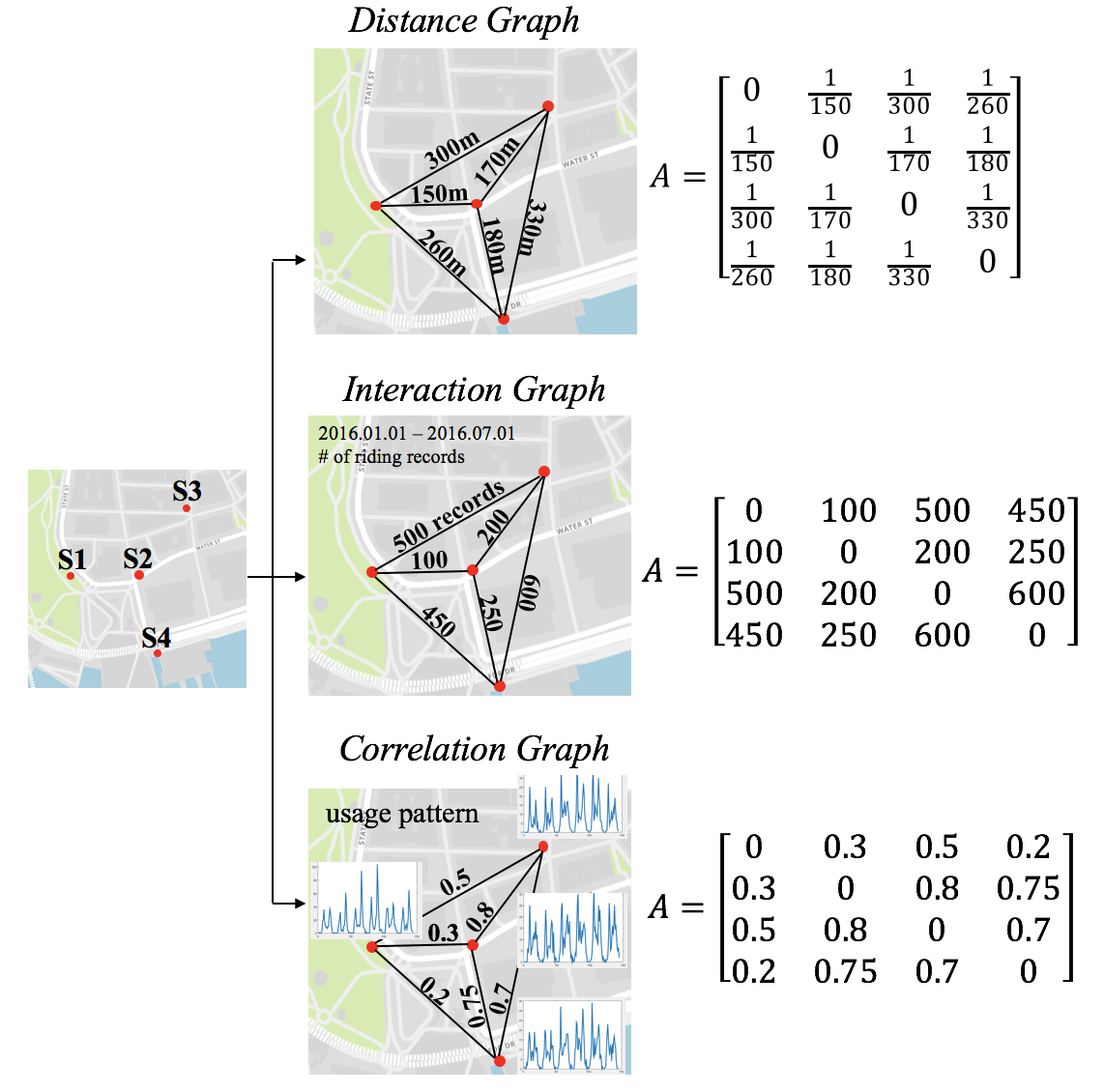}
	\vspace{-2em}
	\caption{Example of different inter-station graphs.}
	\label{fig:graphExample}
\end{figure}

For readers' understanding, Figure~\ref{fig:graphExample} gives a toy example of the above three graph construction methods on four stations. 

\vspace{+.5em}
\noindent \textbf{(ii) Multi-graph Convolution}

To fully exploit different inter-station graphs that contain heterogeneous useful spatial correlation information, we propose a novel multi-graph convolutional layer in our neural network model. It is able to conduct graph convolution on different kinds of graphs by merging them together first. There are  two major steps of multi-graph convolution part: \textit{graph fusion} and \textit{graph convolution}.

\textbf{Graph fusion}: The graph fusion step merges different graphs into one fused graph. We combine different graphs by the weighted summing their adjacency matrices at the element level.  Since the adjacency matrices' value of different graphs may vary a lot (see Figure~\ref{fig:graphExample} for examples), we first normalize the adjacency matrix $A$ for each graph.
$$ A{'}=D^{-1}A+I $$
where $D$ is :
\begin{small}
	$$
	D =
	\left( \begin{array}{cccc}
	\sum_{j=0}^{N-1}A_{0,j}&0&\ldots&0\\
	0                     & \sum_{j=0}^{N-1}A_{1,j} & \ldots  & 0           \\
	\vdots                & \vdots                & \ddots  & \vdots      \\
	0                     & 0                     & \ldots  & \sum_{j=0}^{N-1}A_{N-1,j} 
	\end{array} \right)
	$$
\end{small}

The resultant $A{'}$ is the normalized adjacency matrix with \textit{self loop}. Self-loop can maintain the information of the target station itself in the convolution part, which is a required design strategy in graph convolutional neural networks.

To keep the fusion result normalized after the weighted sum operation, we further add a softmax operation to the weight matrix. Suppose we have $N$ graphs to blend together, we can denote the graph fusion process as:
$$ W_1', W_2', ..., W_N' = Softmax(W_1, W_2,...,W_N) $$
$$ A_i'=D_i^{-1}A_i+I \quad (1 \le i \le N) $$
$$ F = \sum_{i=1}^NW_i' \circ A_i' $$
where $\circ$ is the element-wise product, $F$ is the graph fusion result which will be used in the graph convolution part.

\textbf{Graph convolution}: Based on the graph fusion result $F$, we perform the graph convolution as:
$$ H_0^{t'}=(I^{t'},O^{t'}),\ t' \in [0, t-1]$$
$$H_1^{t'} = F * W_c * H_0^{t'}$$
where $W_c$ is the convolution weight matrix, $H_0^{t'}$ is the bike flow at time $t'$ (inflow $I^{t'}$ and outflow $O^{t'}$). We take $H_1^{t'}$ as the convolution result, and then use $H_1^{t'}$ as the input of the next prediction network.\footnote{We can stack several graph convolutional layers in our neural network model. In this work, for brevity, we just use one graph convolutional layer, and our experiment shows that even with only one layer, our method can already outperform traditional methods significantly in prediction accuracy.}

The graph convolution operation is performed with the filter matrix $W_c$ over the whole bike flow matrix $H_0^{t'}$ at time $t'$. It is worth noting that, although the size of the filter $W_c$ is equal to the size of $H_0^{t'}$, it is still (roughly) a local convolution at the corresponding station due to the existence of the inter-station graph $F$. The reason is that the graph is not fully connected if we build the interaction graph (i.e., two stations are too far from each other to have interacted rides), or most weights of the edge is near zero if we build the distance graph. In a word, many entries in $F$ will be very close to zero. Then, the difficulty of tuning the weight matrix $W_c$ is reduced in the network parameter training process because the initial values of part of $W_c$ will be zero or near zero after multiplied with $F$. 

\begin{figure}
	\centering
	\includegraphics[width=1\linewidth]{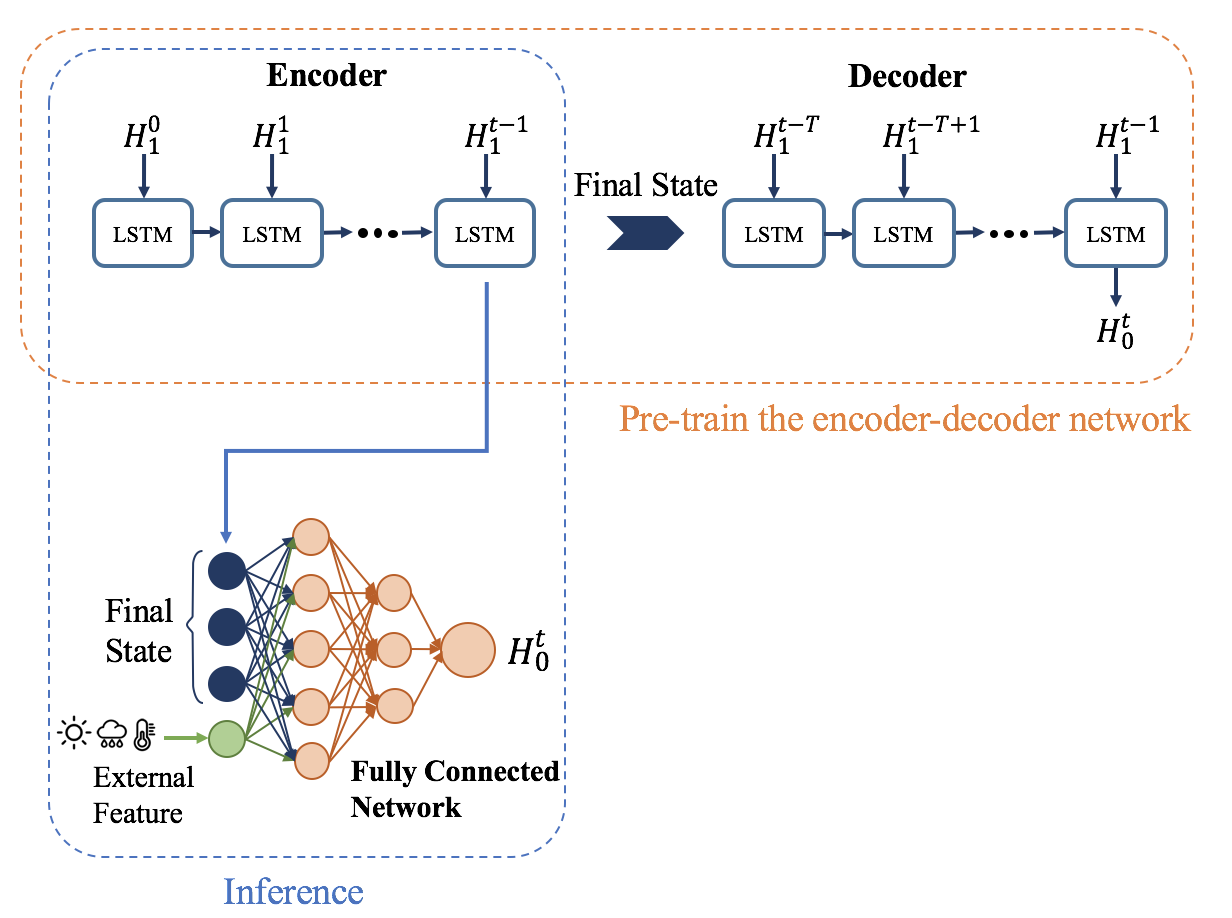}
	\vspace{-2em}
	\caption{Structure of Prediction Network}
	\label{fig:two}
\end{figure}

\vspace{+.5em}
\noindent \textbf{(iii) Prediction Network}

The structure of our prediction network is shown in Figure ~\ref{fig:two}. First, we want to highlight that while multi-graph convolutional layers are able to capture diverse spatial correlations between stations, the temporal patterns in a station's historical ride records have not been exploited yet. Hence, we include the LSTM layers in our neural network model to catch the temporal patterns after the multi-graph convolution. Moreover, we leverage an \textit{encoder-decoder structure} along with LSTM layers for two reasons. First, encoder-decoder structure has been verified very effective in spatio-temporal prediction tasks, and now is one of the most widely used neural network prediction structures \cite{sutskever2014sequence}. Second, with the encoder-decoder structure, in fact we can also infer the prediction uncertainty, or \textit{confidence interval}, which we will elaborate later.

As shown in Figure~\ref{fig:two}, the details of the encoder-decoder structure is as follows. The encoder network takes the multi-graph convolutional result sequence $[H_1^0,H_1^1,...,H_1^{t-1}]$ as input and encodes the temporal pattern into the final state of LSTM cell after the rolling process. The decoder network takes the encoder's final state as the initial state and the multi-graph convolutional result sequence $[H_1^{t-T},H_1^{t-T+1},...,H_1^{t-1}]$ as input. The output of the decoder is $H_0^t$ which is our prediction target. We can set $T$ to a small value (e.g., half of $t$) which means that the decoder can predict the future bike flow based on a short period of history data and the encoder's final state. This implies that the encoder's final state provides important information for the predicting process. After training the encoder-decoder structure, we input the encoder network's final state, combined with some external context features (e.g., temperature, wind speed, weekday/weekend~\cite{zhang2017deep}) to a fully connected network (lower part of Figure~\ref{fig:two}) for predicting the bike flow in the next time $H_0^t$.

\begin{figure}
	\centering
	\includegraphics[width=1\linewidth]{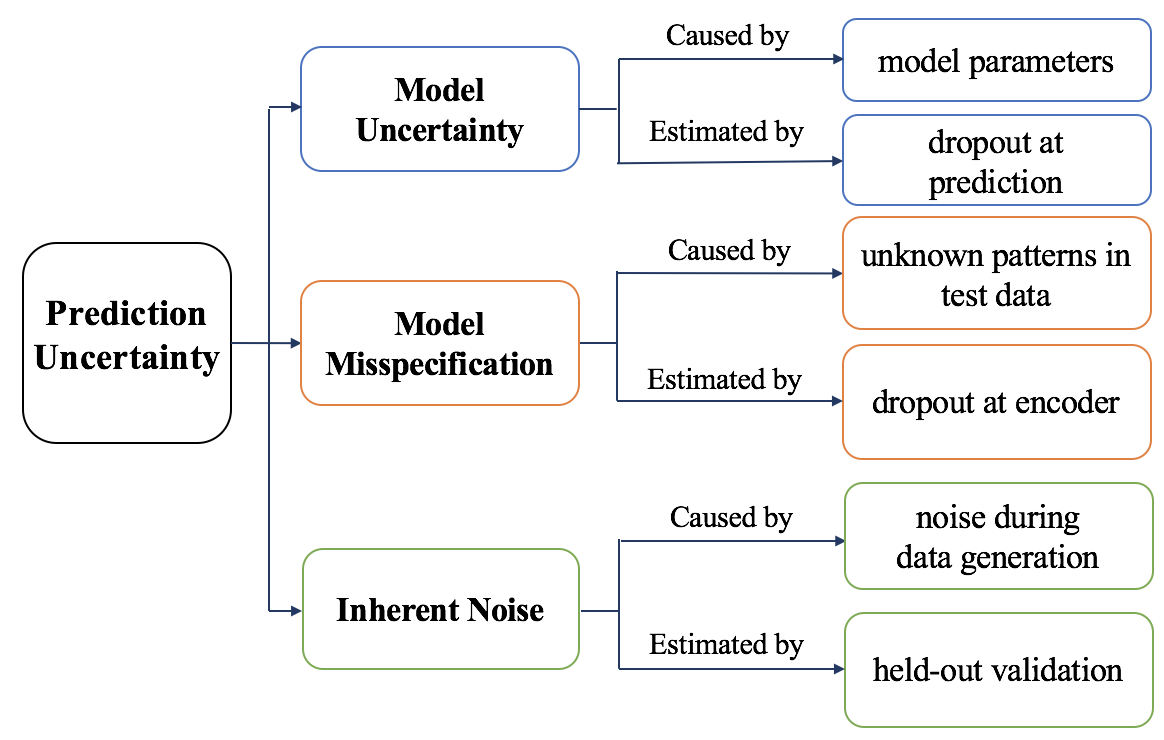}
	\caption{Component of prediction uncertainty}
	\label{fig:uncertaintyEstimation}
\end{figure}

\textbf{Confidence Estimation:} Next, we elaborate how to infer the uncertainty of our prediction, i.e., confidence interval based on the encoder-decoder prediction network. According to literature, the uncertainty of the prediction result can be divided into three parts: \textit{model uncertainty}, \textit{model misspecification}, and \textit{inherent noise}~\cite{zhu2017deep} (Figure~\ref{fig:uncertaintyEstimation}). 
\begin{enumerate}
	\item \textit{Model uncertainty}, also called  epistemic uncertainty, is the variance caused by the uncertainty of the trained model parameters. This uncertainty can often be reduced with the increase of training dataset. 
	\item \textit{Model misspecification} appears when the test dataset contains some different patterns from the training dataset. Such uncertainty is more common when the test data is sampled from a different distribution of the training data. In bike flow prediction, this is non-ignorable as in the future, there may happen some special events that have never happened before, thus leading to the model misspecification issue.
	\item  \textit{Inherent noise} emerges when the data is originally generated, which is irreducible in practice.
\end{enumerate}

To estimate the model uncertainty and misspecification, previous works have pointed out that the precise calculation is hard due to the non-linear property of neural networks, while dropout techniques can provide a useful approximation \cite{zhu2017deep,gal2016dropout}. That is, given an input sequence of the bike flow matrix, during the inference, we perform dropouts in the encoder network (i.e. LSTM units) to get various embeddings for the input (i.e., final state of LSTM units). The variance in such embeddings can represent the model misspecification. At the same time, we also conduct dropouts in the fully-connected prediction network to approximate the model uncertainty. In a word, by performing dropouts in both encoder and fully connected networks for an input, we can estimate the model uncertainty and misspecification by calculating the variance of output prediction results under different dropout trials. Note that the variational dropout is used for LSTM as it has been verified more effective for recurrent neural networks~\cite{gal2016theoretically}. For the inherent noise, an easy way to estimate it is using a held-out validation dataset to calculate the prediction error variance, which can be proved to be an asymptotically unbiased estimation on the inherent noise level~\cite{zhu2017deep}.

Finally, after obtaining the prediction variance $\sigma_1$ incurred by the model uncertainty and model misspecification with dropout, and variance $\sigma_2$ incurred by inherent noise with a held-out validation dataset, we can infer the confidence interval as:
$$[y^*-Z_{\frac{\alpha}{2}}*\sqrt{\sigma_1^2+\sigma_2^2}, y^*+Z_{\frac{\alpha}{2}}*\sqrt{\sigma_1^2+\sigma_2^2}]$$
where $y^*$ is the prediction result for a certain station at a time slot, and $Z_{\frac{\alpha}{2}}$ is the $(1-\alpha)$ confidence interval of standard normal distribution.

\section{Evaluation}

In this section, we evaluate our multi-graph convolutional network method with real bike sharing datasets. We will first introduce the dataset, experiment settings, and then illustrate our experiments results comprehensively.

\subsection{Experiment Setting}

\textbf{Datasets:} We used bike flow dataset collected from New York City and Chicago\footnote{NYC bike sharing data: \url{https://www.citibikenyc.com/system-data}, Chicago bike sharing data: \url{https://www.divvybikes.com/system-data}}. The datasets cover a four year time period. All the data are in the forms of riding record containing start station, start time, stop station, stop time and so on. We summarize the dataset statistics in table~\ref{tab:one}. Weather data comes from the NCEI website\footnote{\url{https://www.ncdc.noaa.gov/data-access}} (National Centers for Environmenttal Information).

To set the training-validation-test data split, we choose the last 80 days in each city as test data, the 40 days before the test data are validation data, and all of the data before validate data are training data. The prediction granularity is set to one hour.

\begin{table}[t]
	\caption{Dataset statistics}
	\label{tab:one}
	\vspace{-1em}
	\begin{minipage}{\columnwidth}
		\begin{center}
			\begin{tabular}{lll}
				\toprule
				& New York City & Chicago\\
				\midrule
				Time span     & 2013.07-2017.09 & 2013.06-2017.12\\
				\#Riding records
				& 49,669,470 & 13,826,196\\
				\#Stations     & 827 & 586\\
				\bottomrule
			\end{tabular}
		\end{center}
	\end{minipage}
\end{table}%

\vspace{+.5em}
\noindent
\textbf{Network Implementation and Parameters:} The encoder and decoder implemented in the experiment contain one layer of LSTM and 64 hidden units. The fully connected prediction network contains 4 layers including the input and output layer. We choose the optimization algorithm as ADAM and the learning rate is set to 0.001\%~\cite{kingma2014adam}. 

In the encoder-decoder structure of the prediction network, we set $T=3$ in the decoder (refer to Figure~\ref{fig:two}). We use the past 6-hour history data to predict the bike flow in the next one hour. For the confidence computation, we compute the 95\% confidence interval for the prediction result. For brevity, we only report the prediction results of bike inflow in this section, while the bike outflow results are very similar. Note that all of the above network implementation and parameter settings are chosen as they can perform well on the 40-day validation data, while the reported results in the next subsections are based on the 80-day test data.

\vspace{+.5em}
\noindent
\textbf{Baselines:} We compare our multi-graph convolutional network model with the following baselines:
\begin{itemize}
	\item \emph{ARIMA} \cite{williams2003modeling}: Auto-Regressive Integrated Moving Average is a widely used time series prediction model. 
	
	\item \emph{SARIMA} \cite{williams2003modeling}: The seasonal version of ARIMA. 
	
	\item \emph{LSTM} \cite{tian2015predicting}: Recent studies in traffic flow prediction, such as \cite{tian2015predicting}, adopted the long short-term memory (LSTM) recurrent neural network model and verified its effectiveness.
\end{itemize}

%
%

\subsection{Experiment Results}

\begin{table}[t]
	\caption{Prediction error in New York City and Chicago. Top stations are ranked by each station's total sum of bike flows in the historical ride records.}
	\vspace{-1em}
	\label{tab:two}
	\begin{minipage}{\columnwidth}
		\begin{center}
			\begin{tabular}{lccc}
				\multicolumn{4}{c}{New York City} \\
				\toprule
				& \textit{Top 5 stations} & \textit{Top 10 stations}& \textit{Average}\\
				\midrule
				\emph{ARIMA} & 11.329 &	9.545 &	8.049 \\
				\emph{SARIMA} & 8.677 &	7.363 &	6.521 \\
				\emph{LSTM} & 6.802 &	5.981 &	5.345\\
				\emph{Multi-graph} & \textbf{4.745} &	\textbf{4.473} &	\textbf{4.003}\\
				\bottomrule
				& & & \\
				\multicolumn{4}{c}{Chicago} \\
				\toprule
				& \textit{Top 5 stations}& \textit{Top 10 stations}& \textit{Average}\\
				\midrule  
				\emph{ARIMA} & 9.853 &	8.535 &	6.163  \\
				\emph{SARIMA} & 6.797 &	6.175 &	4.608 \\
				\emph{LSTM} & 6.231 &	5.853 &	4.405\\
				\emph{Multi-graph} & \textbf{5.177} &	\textbf{4.930} &	\textbf{3.658} \\
				\bottomrule
			\end{tabular}
		\end{center}
	\end{minipage}
\end{table}%

\textbf{Prediction error}: We use RMSE (root mean square error) to measure the prediction error. Table ~\ref{tab:two} shows our evaluation results. In general, among various baselines, LSTM can perform the best. However, our multi-graph convolutional method can still beat LSTM significantly by reducing the average prediction error by 25.1\% and 17.0\% in New York City and Chicago, respectively. 

In addition to the average station-level prediction error,  we investigate the prediction results for those stations with the highest usages in both cities. The prediction accuracy of these busy stations may be more important since most of the over-demand issues (`no bikes to use' or `no docks to return a bike')  could happen in those stations \cite{chen2016dynamic}. We thus select the top 5 and 10 busy stations to study the results in both cities. We observe that our multi-graph method can also consistently get better results for the busy stations. For example, for the top 5 and 10 busy stations in New York City, our method can outperform LSTM by 30.2\% and 25.2\%, respectively. This further verifies the practicability of our proposed method in real-life bike sharing system management.


\begin{figure}[t]
	\centering                                                          
	\includegraphics[width=1\linewidth]{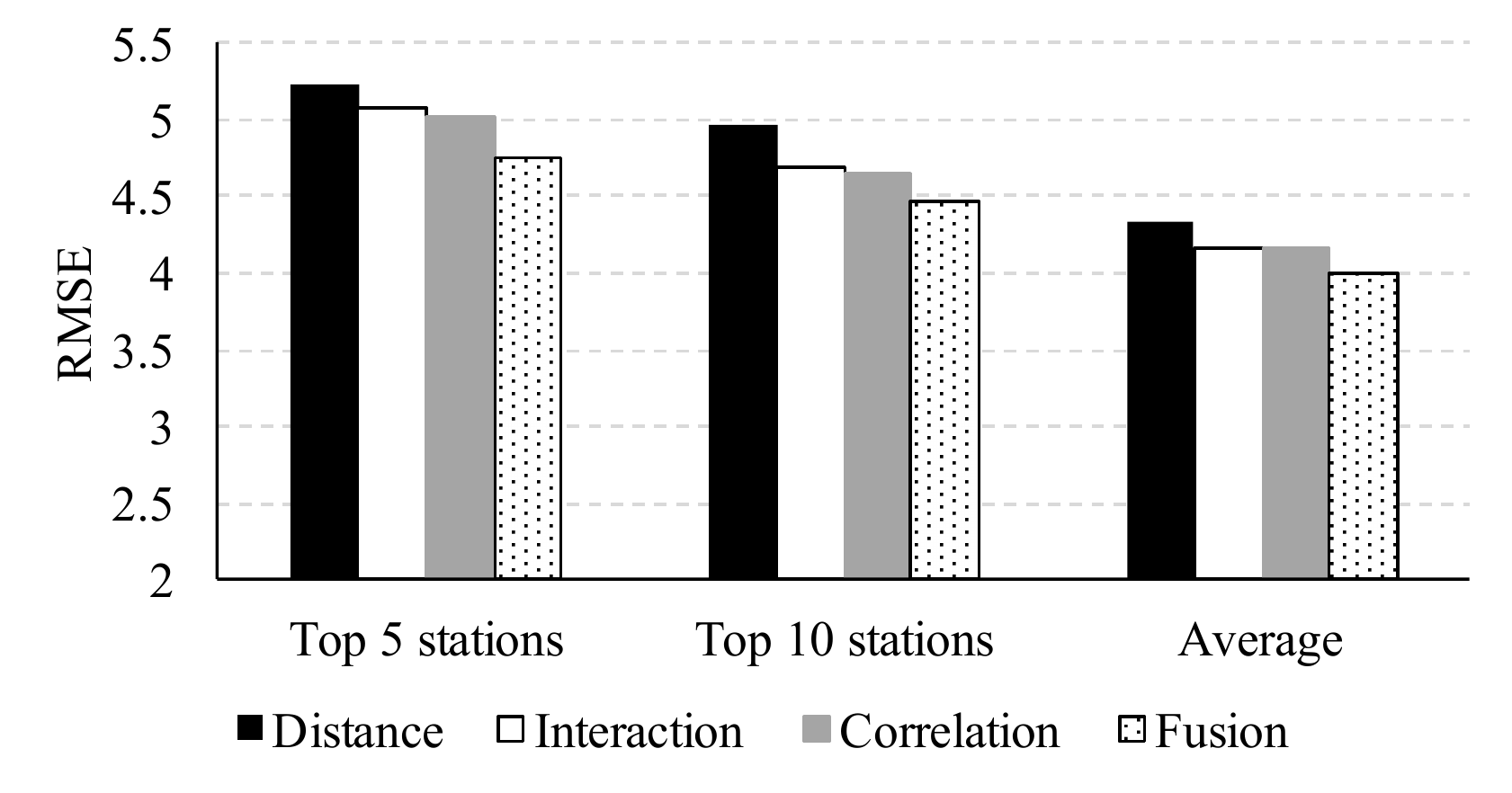}
	\vspace{-1.5em}
	\caption{Comparison of multi-graph and single-graph convolutional models (New York City).}
	\label{fig:multi_vs_single_nyc}
\end{figure}

\begin{figure}[t]
	\centering                                                          
	\includegraphics[width=1\linewidth]{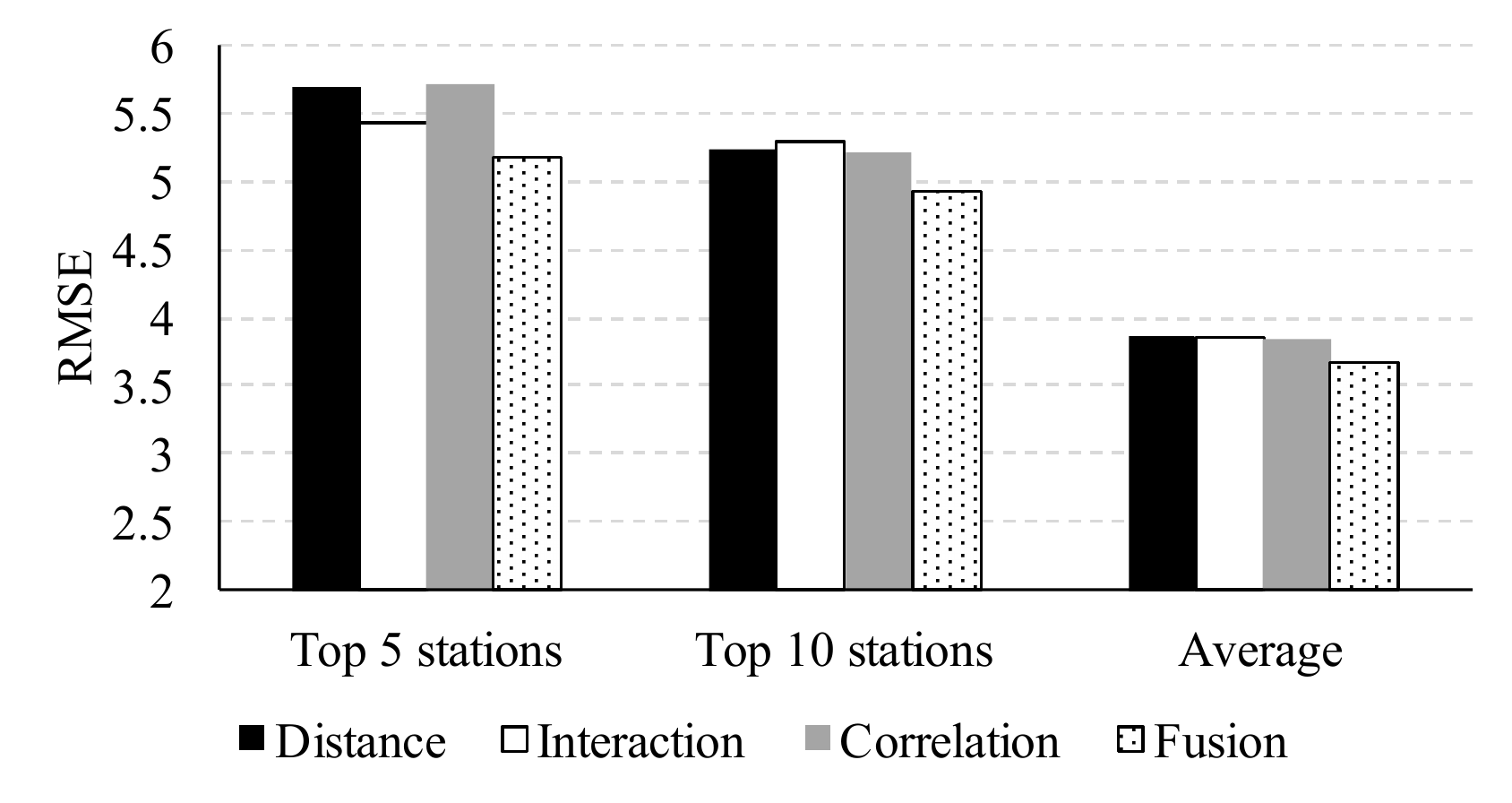}
	\vspace{-1.5em}
	\caption{Comparison of multi-graph and single-graph convolutional models (Chicago).}
	\label{fig:multi_vs_single_chicago}
\end{figure}

\vspace{+.5em}
\noindent
\textbf{Effectiveness of multi-graph fusion}: Now we verify that our multi-graph fusion strategy can actually bring benefits to the prediction model. Figure~\ref{fig:multi_vs_single_nyc} and \ref{fig:multi_vs_single_chicago}  show the results when we only use a single graph (\textit{distance}, \textit{interaction} or \textit{correlation} graph) for prediction in New York City and Chicago. Compared to the single-graph convolutional methods, our multi-graph convolutional method can perform consistently better. For example, for the top 5 busy stations in New York City, the multi-graph model can outperform the single-graph models by reducing error 5.6--9.2\%.

Among the single-graph methods, we can observe that which graph model performs better is dependent on the test stations. For example, for the top 5 busy stations in Chicago, the best single-graph convolutional model is `interaction graph'. However, when we evaluate on the top 10 busy stations, it performs worse than the other two single graphs, `distance graph' and `correlation graph'. So, if we use the single-graph method, how to choose the graph would be very hard and tricky. In comparison, with our proposed multi-graph method, we do not need to bother selecting which single graph representation of the bike sharing system. Our proposed fusion method can automatically and adaptively extract useful information from all of the single graphs and then achieve better prediction performance.




\vspace{+.5em}
\noindent
\textbf{Confidence interval estimation}: To evaluate whether our confidence interval estimation is accurate, we calculate the ratio that the real bike flow values fall into our estimated confidence interval. If the ratio is close to 95\%, then it means that the estimation is accurate. Table ~\ref{tab:three} shows the result of the confidence interval computation. In addition to our method which considers three components of uncertainty (\textit{model uncertainty}, \textit{model misspecification}, and \textit{inherent noise}), we also test the confidence estimation result if we only use dropout or validation variance. We find that our method can achieve the closest value toward 95\%, verifying that both dropout and validation variance are effective in the uncertainty estimation.

\begin{table}[t]
	\caption{The ratio of actual bike flow values falling into the estimated confidence interval (Chicago).}
	\label{tab:three}
	\vspace{-1em}
	\begin{center}
		\begin{tabular}{lc}
			\toprule
			\textit{Method} & \textit{Confidence}\\
			dropout (model uncertainty + misspecification) & 0.486 \\
			validation variance (inherent noise) & 0.869 \\
			our method & 0.933 \\
			\bottomrule
		\end{tabular}
	\end{center}
\end{table}

\begin{figure*}[t]
	\centering                                                          
	\includegraphics[width=.8\linewidth]{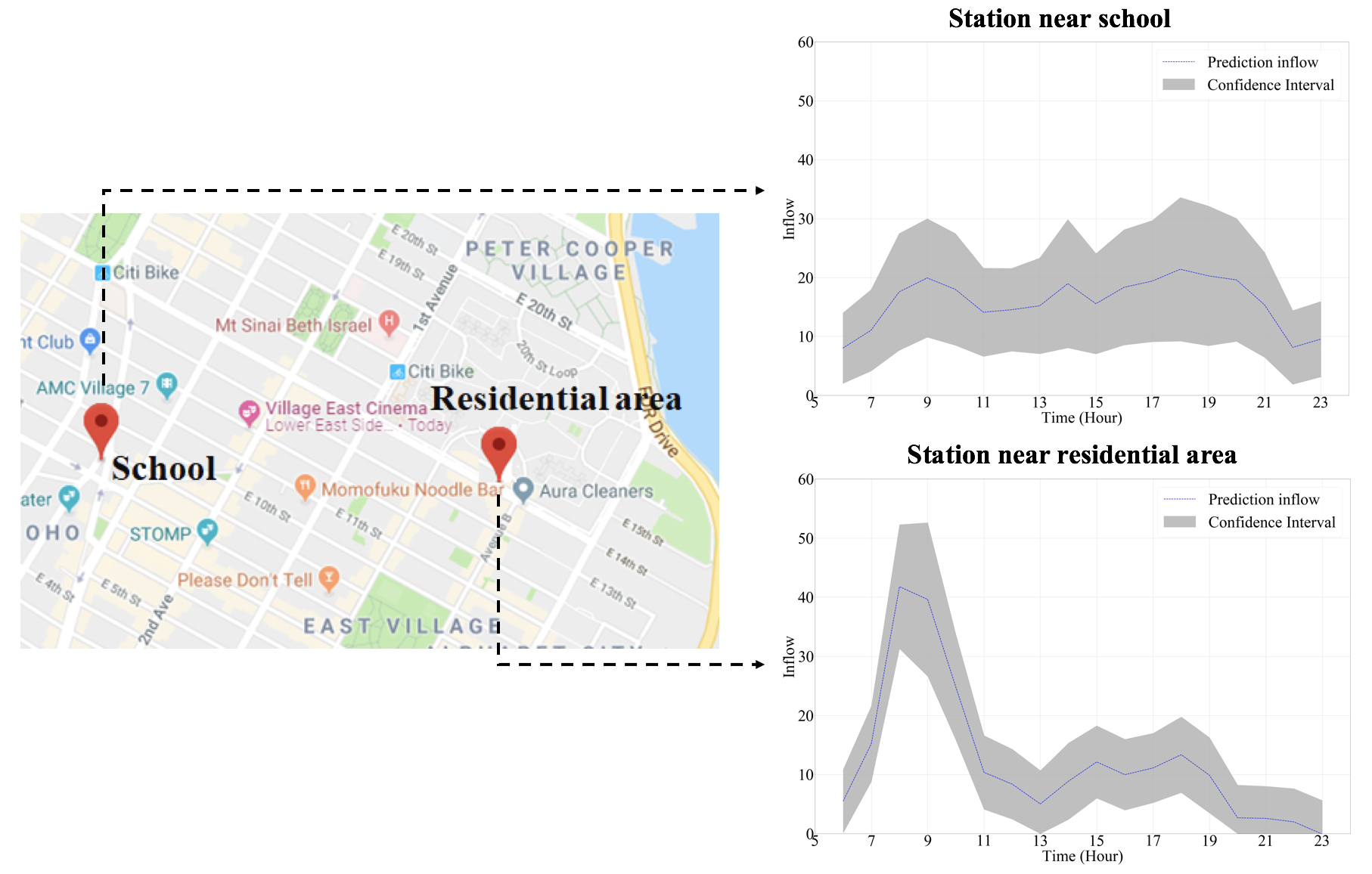}
	\vspace{-1em}
	\caption{Case study of two bike stations with different estimated confidence intervals (New York City).}
	\label{fig:analysisUncertainty}
\end{figure*}


In reality, the confidence interval estimation can provide richer information than only a value estimation. For example, if a station usually has a larger confidence interval in flow prediction, it implies that the station may be intrinsically hard to predict. Then, by studying the stations with larger confidence intervals, we may be able to identify key factors impacting the station predictability, which might further guide us to design a more effective prediction model. Take the two stations in Figure ~\ref{fig:analysisUncertainty} as an example, the two stations have similar daily inflow but perform quite different in the estimated confidence intervals. More specifically, the station near school has a very complicated usage pattern, leading to a large estimated confidence interval; in comparison, the station near residential area performs more regularly, even though it has a peak usage pattern around 8:00 a.m. From this result, we may infer that there is much more space for the station near school to improve its prediction accuracy. Then, the bike sharing system manager can devote more efforts to such hard-to-predict stations and explore whether more factors can be incorporated into the prediction model to increase these stations' prediction performance.




\begin{figure}[t]
	\centering                                                          
	\includegraphics[width=.9\linewidth]{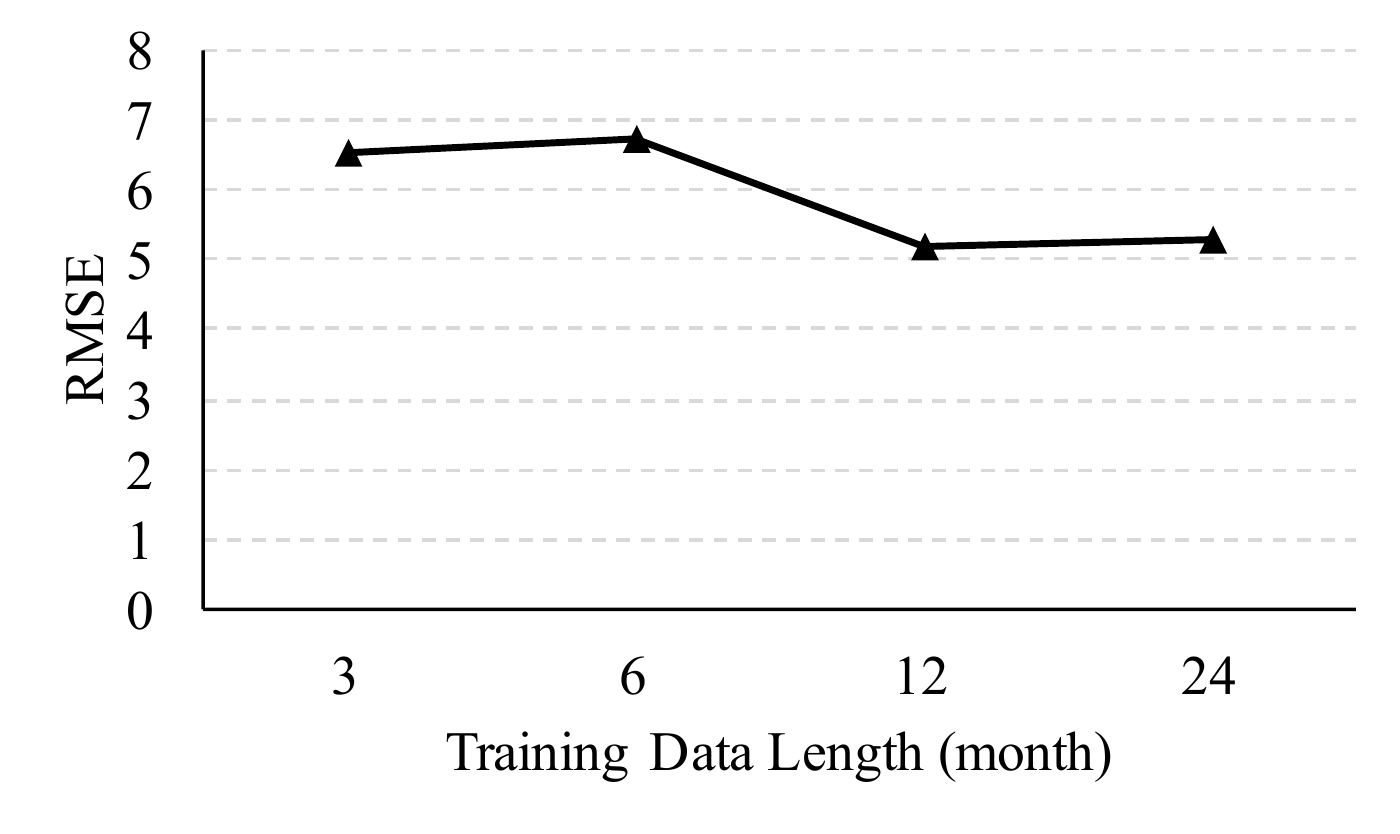}
	\vspace{-1.5em}
	\caption{Tuning the length of training data (Chicago).}
	\label{fig:tune_train_len}
\end{figure}

\vspace{+.5em}
\noindent
\textbf{Tuning the training data length}: In our method, the training data has two major roles: (i) building the inter-station graphs such as interaction and correlation graphs; (ii) training the whole multi-graph neural network model parameters. Hence, if the training data length is too short, the prediction results may not be satisfactory. 

To test how long training data is needed to achieve a good performance, we vary the length of training data and see how the prediction error changes. Figure~\ref{fig:tune_train_len} shows the results in Chicago. We can find that if the training data length is shorter than 6 months, our model does not perform very well. By increasing the length of training data beyond 12 months, the prediction error reduces significantly. With these results, we suggest that for a robust prediction accuracy, the training data is desired to be last for at least one year. While some cities may not have enough historical records if they just start the bike sharing systems, we plan to study how to address the cold-starting problem in the future work, e.g., with cross-city knowledge transfer learning methods~\cite{wang2018crowd}.

\begin{figure}[t]
	\centering                                                          
	\includegraphics[width=.9\linewidth]{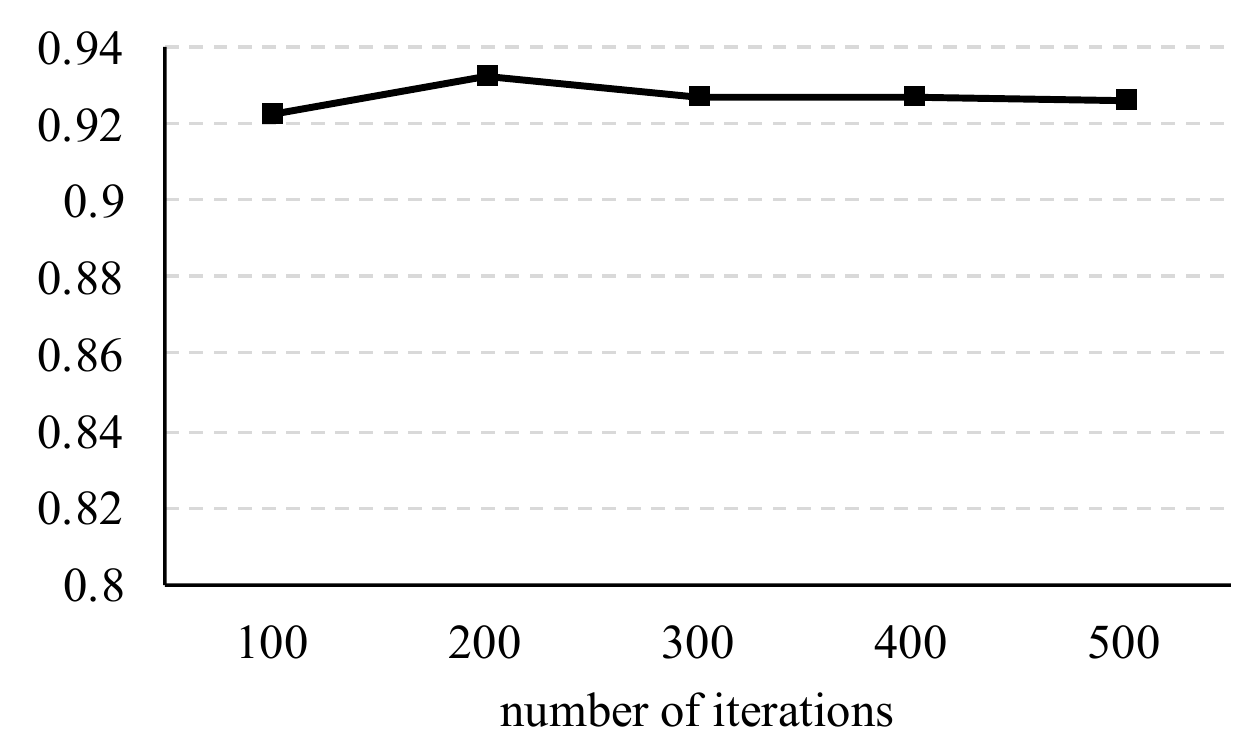}
	\vspace{-1.5em}
	\caption{Actual coverage ratio of the estimated confidence interval by varying the number of dropout iterations (Chicago).}
	\label{fig:uncertainty_iter}
\end{figure}

\vspace{+.5em}
\noindent
\textbf{Tuning the number of dropout iterations}: When calculating the confidence interval, we need to simulate several iterations of dropouts so as to estimate the model uncertainty and model misspecification. Here, we test which number of iterations is needed for obtaining a robust confidence interval estimation. As shown in Figure~\ref{fig:uncertainty_iter}, we can find that the coverage ratio of actual bike flow values by our estimated confidence does not change significantly when we conduct 100 to 500 iterations, especially after 300 iterations. Hence, we think that several hundred of iterations should be enough for the confidence interval estimation of the station-level bike flow prediction.

\vspace{+.5em}
\noindent
\textbf{Computation efficiency}: Our experiment runs in a Windows server with \textit{CPU: Intel Xeon E5-2690, Memory: 56 GB, GPU: Nvidia Telsa K80}. The training time needs about 2 to 3 hours, while the inference just takes a few seconds. Since the training process is an offline process, this running efficiency is enough for real-life bike flow prediction systems.





\section{Conclusion and Future work}

In this paper, we propose a new multi-graph convolutional neural network model to predict station-level bike flow in a bike sharing system. There are two novel aspects of our model. The first aspect is the multi-graph convolution part which utilizes the graph information in flow prediction. More specifically, we design three heterogeneous inter-station graphs to represent a bike sharing system, namely \textit{distance}, \textit{interaction}, and \textit{correlation} graphs; a fusion method is then proposed to conduct the graph convolution operation on the three graphs simultaneously.  The second aspect is the uncertainty computation part that is able to infer the confidence interval for our prediction. The confidence interval estimation employs the dropout technique to obtain a robust estimation interval.

As the pioneering effort to employ the graph convolutional networks on bike flow prediction, there are still many issues to investigate in the future.

\textit{Extending usage scenarios}. There are many other urban traffic systems similar to bike sharing, such as subway. We are now working on extending our bike flow prediction model to a more general urban traffic prediction methodology.

\textit{Anomaly detection}. With confidence interval estimation, another important usage is the anomaly detection. That is, if we detect an irregular large uncertainty for a station at some time slots (compared to average), then it is probably that some abnormal events happen around the station. We will test how such anomaly detection works in real-life bike sharing systems.

\textit{Addressing cold-start problems.} As shown in our experiments, our current model needs more than one-year historical bike flow records to obtain a good prediction accuracy. One of the important future issues is to reduce the length of required training data length, so as to address the cold-start problem of the bike flow prediction.

\textit{Improving network structure}. In this work, we use LSTM as the basic units to capture temporal patterns in bike flow. Very recent studies \cite{vaswani2017attention,lianggeoman} have indicated that attention units may be potentially more effective. Hence, we will study whether replacing LSTM with attention can further boost the prediction performance.



\bibliographystyle{plain}
\bibliography{paperRef}
\end{document}